\documentclass[letterpaper]{www2005-submission}

\usepackage[T1]{fontenc}
\usepackage{ae,aecompl}
\usepackage[latin1]{inputenc}
\usepackage{array}
\usepackage{float}
\usepackage{amsmath}
\usepackage{graphicx}
\usepackage{amssymb}

\makeatletter

\providecommand{\tabularnewline}{\\} \floatstyle{ruled}
\newfloat{algorithm}{tbp}{loa}
\floatname{algorithm}{Algorithm}

\usepackage{verbatim}

\def\sharedaffiliation{%
\end{tabular}
\begin{tabular}{c}
}

\makeatother
\begin{document}

\title{Emergence of Spontaneous Order Through Neighborhood Formation in
Peer-to-Peer Recommender Systems}

\author{\numberofauthors{3}\\
\alignauthor Ernesto Diaz-Aviles\\
 \alignauthor Lars Schmidt-Thieme\\
 \alignauthor Cai-Nicolas Ziegler\\
\sharedaffiliation\\
 \affaddr{Institut für Informatik, Universität Freiburg}\\
\affaddr{Georges-Köhler-Allee, Gebäude 51}\\
\affaddr{79110 Freiburg i.Br., Germany}\\
\hfill{}\\
\small{\{diaz,lst,cziegler\}@informatik.uni-freiburg.de}\\
} \maketitle

\begin{abstract}
The advent of the Semantic Web necessitates paradigm shifts away from
centralized client/server architectures towards decentralization and
peer-to-peer computation, making the existence of central authorities
superfluous and even impossible. At the same time, recommender systems
are gaining considerable impact in e-commerce, providing people with
recommendations that are personalized and tailored to their very needs.
These recommender systems have traditionally been deployed with stark
centralized scenarios in mind, operating in closed communities detached
from their host network's outer perimeter. We aim at marrying these
two worlds, i.e., decentralized peer-to-peer computing and recommender
systems, in one agent-based framework. Our architecture features an
epidemic-style protocol maintaining neighborhoods of like-minded peers
in a robust, self-organizing fashion. In order to demonstrate our
architecture's ability to retain scalability, robustness \emph{and}
to allow for convergence towards high-quality recommendations, we
conduct offline experiments on top of the popular MovieLens dataset.
\end{abstract}




\section{Introduction}

Today's most successful recommender systems are part of e-commerce
infrastructures, and they share a common characteristic: they are
all based on a client/server architecture (or an \emph{N}-tier
variation of it), where the user profile information and
recommendation engine are \emph{centralized}.

However, the Semantic Web vision \cite{TBLEE2001SW} that we share
is more likely to be based on decentralized architectures, like
the ones provided by peer-to-peer (P2P) overlay networks, where
\emph{agents} would interact via free information exchange or
trading. We present an alternative to centralized collaborative
filtering, exploiting
the advantages of peer-to-peer networks.\\

\begin{figure}[H]

\begin{center}\includegraphics[%
 scale=0.13]{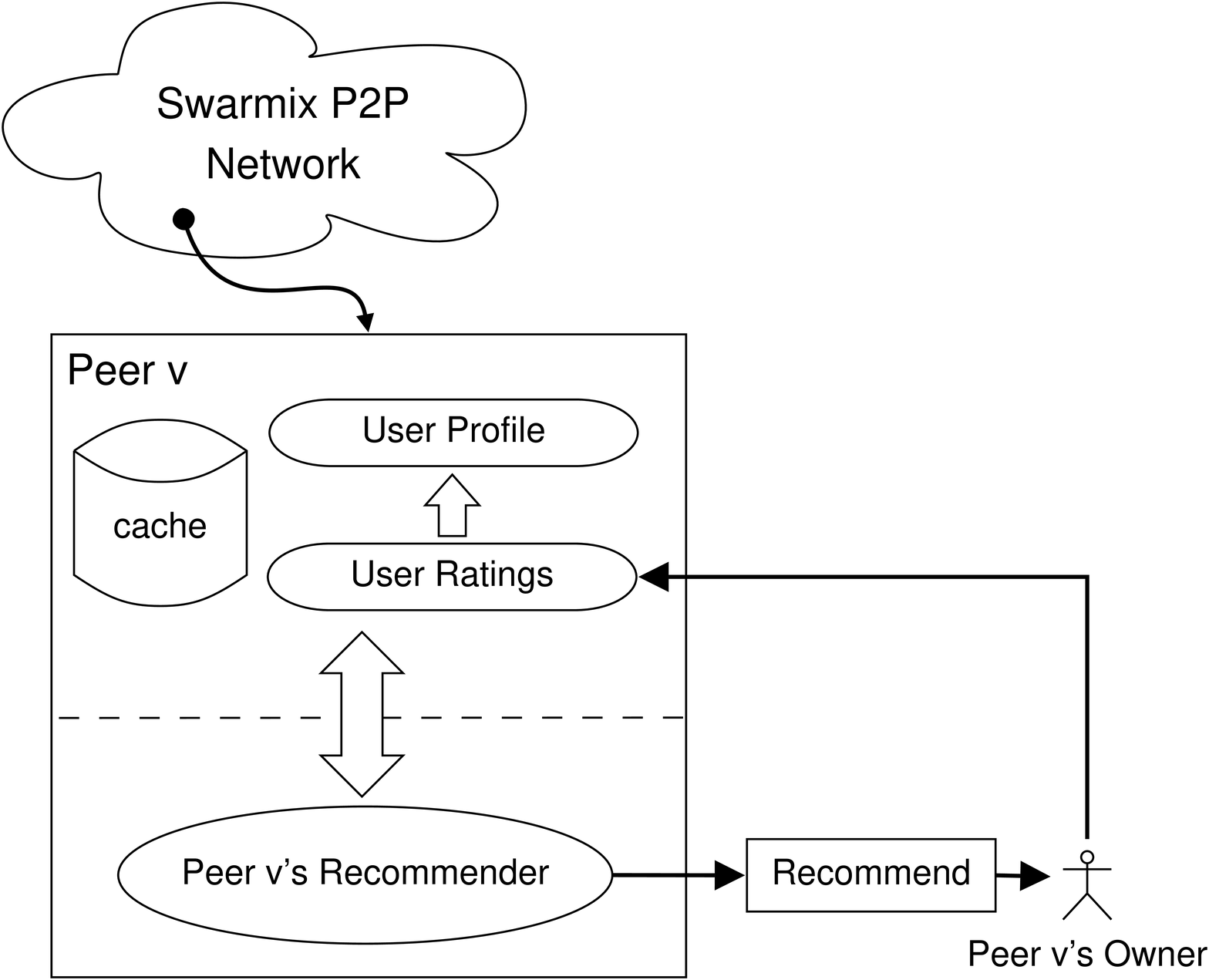}\end{center}

\caption{Overview of the Swarmix distributed recommender system
architecture.\label{fig:Swarmix Overview DISTSYS}}

\end{figure}

We introduce Swarmix, a distributed architecture (Figure
\ref{fig:Swarmix Overview DISTSYS}) whose epidemic-style protocol
is responsible for the overlay P2P network construction and
maintenance. The protocol is able to associate each peer $v$ with
a fixed number of highly similar neighbors whose similarity with
respect to $v$ improves during the perpetual execution of the
protocol. Each peer $v$ runs a recommender system locally and is
in control of its profile and ratings; $v$'s recommendations are
computed using only its peers; which requires no global knowledge
of the network or access to a central server responsible for
storing or computation.

The rest of the paper is organized as follows. In Section 2, we present
a general model shared by epidemic-style protocols based on a push-pull
mechanism. In Section 3, we introduce the Swarmix protocol at the
core of our architecture. The distributed recommender system implementation
is presented in Section 4. In Section 5, we present the experimental
setup and evaluation metric used. In Section 6, we report our experimental
results. In Section 7, we point to some related work. Finally, Section
8 presents our conclusions and future research.

\section{The General Epidemic \protect \\
Push-Pull Protocol (GEP3)\label{sec:General-Epidemic-Push-Pull}}

Epidemic-style distributed protocols like the \emph{anti-entropy
push-pull} mechanism \cite{43922}, or the \emph{Newscast} protocol
\cite{newscast03,jelasity02largescale} share the same general
model that we call the \emph{general epidemic push-pull protocol}
(GEP3), which is detailed in this section.

We formalize the task to distribute data in a peer-to-peer network
as follows: Let $V$ be a set representing the \textbf{peers}, $S$
a set representing \textbf{data}. Then each peer $v\in V$ stores

\begin{enumerate}
\item the set of peers $Q_{v}$ it is able to communicate with, called \textbf{neighborhood}%
\footnote{The set $V$ of peers and the neighborhoods $(Q_{v})_{v\in V}$ form
a directed graph called \textbf{network topology.}%
},
\item a set $C_{v}\in{\mathcal{P}}(S)$ of data called the peers \textbf{cache}--where
${\mathcal{P}}(.)$ denotes the power set--, and
\item a \textbf{utility function} $u:{\mathcal{P}}(S)\rightarrow{\mathbb{R}}$
that specifies for each possible state of the cache the utility for
the peer. For simplicity, we will only consider decomposable utility
functions with $u(C)=\sum_{s\in C}u(s)$ in the following.
\end{enumerate}
Initially, each peer may have some data, and new data or new
versions of old data may enter the system through any peer, at any
time. Data is transmitted through the network by exchanging and
merging the caches of two neighboring peers $v$ and $w$ with the
goal to maximize the utility of each peer's cache, conforming to
some constraints as, e.g., a maximal cache size.

As several copies and versions of the same data may pile up during
runtime at each peer, we need some method for duplicate elimination\[
C_{v}^{\prime}:=\text{merge}(C_{v},C_{w})\]
called \textbf{cache merging} in the following. For version handling,
let $\text{id}:S\rightarrow\text{IDs}$ be a function that returns
an ID for each data, and $t:S\rightarrow\text{Time}$ be a function
that returns the \emph{timestamp} of the data's version, then
\textbf{merging by retaining the most recent version} for each
data is implemented as follows:

\[
\text{merge}(C_{v},C_{w}):=\{\textrm{argmax}_{\substack{r\in C_{v}\cup C_{w},\\
id(r)=\text{id}(s)}
}t(r)\:|\: s\in C_{v}\cup C_{w}\}\]

Then, we \textbf{retain a fixed number of most useful data} using
method\[ C_{v}^{\prime\prime}:=\textrm{select}(C_{v}^{\prime},k)\]

\noindent called \textbf{selection function}, implemented as
follows:

\[
\text{select}(C_{v}^{\prime}):=\textrm{argmax}_{s\in C_{v}^{\prime}}^{k}u(s)\]
where $k$ is a fixed integer and $\text{argmax}^{k}$ returns the
$k$ first elements in descending order of its argument.

Finally, the protocol allows to adapt the network topology dynamically
based on local information, accomplished by a function\[
Q_{v}^{\prime}:=\text{neighbors}(Q_{v},C_{v}^{\prime\prime})\]
called \textbf{neighborhood adaptation}. If each data contains information
about the peer at which it entered the network, modeled by a function
$\text{src}:S\rightarrow V$, \textbf{neighborhood adaptation by retaining
a fixed number of most useful peers} is implemented as\[
\text{neighbors}(Q_{v},C_{v}^{\prime\prime}):=\text{src}(\textrm{argmax}_{s\in C_{v}^{\prime\prime}}^{l}u(s))\]
where $l$ is a fixed integer. If \textbf{no neighborhood adaptation}
should be used, i.e., neighborhoods remain constant during runtime,
then $\text{neighbors}(Q_{v},C_{v}^{\prime\prime}):=Q_{v}$.

Note that if each data item (and all its versions) can enter the network
only at a single peer, and if there is exactly one such data item
per peer, then the data ID coincides with the data source ($\text{id}=\text{src}$).

The \textbf{general epidemic push-pull protocol} \textbf{(GEP3)} ran
by each peer $v\in V$ is outlined in Algorithm \ref{alg:General-epidemic-push-pull-Alg}.
Once each $v\in V$ has performed one run of Algorithm \ref{alg:General-epidemic-push-pull-Alg},
a \emph{cycle} has been completed.

\begin{algorithm}

\caption{General epidemic push-pull protocol (GEP3)\label{alg:General-epidemic-push-pull-Alg}}

\textbf{do} forever\vspace*{-0.2cm}

\begin{enumerate}
\item $w:=\textrm{getRandomPeer}(Q_{v})$;
\item \textbf{push} $C_{v}$ to $w$;
\item \textbf{pull} $C_{w}$ from $w$;
\item $C_{v}^{\prime}:=\text{merge}(C_{v},C_{w})$;
\item $C_{v}^{\prime\prime}:=\textrm{select}(C_{v}^{\prime},k)$;
\item $Q_{v}^{\prime}:=\text{neighbors}(Q_{v},C_{v}^{\prime\prime})$;\end{enumerate}

\end{algorithm}

Steps 4 to 6 are executed separately for peers $v$ and $w$ and in
general may give different results, e.g., due to different utility
functions used for selecting the most useful data. We say the GEP3
is \textbf{symmetric} if steps 4 to 6 are guaranteed to return the
same results for both $v$ and $w$.

Note that the three main steps, merging, selection, and neighborhood
adaptation, are most commonly implemented using some sort of weight
function that defines an ordering on the data or peers, but these
weight functions do not have to be the same. Especially if versions
are present, it makes more sense to use the timestamps' 'freshness'
for merging, but the utility for selection and neighborhood adaption.

In the rest of this section, we briefly explore the matching characteristics
between the anti-entropy push-pull mechanism, the Newscast protocol,
and the general protocol presented above.

\begin{figure}

\begin{center}\includegraphics[%
 scale=0.18]{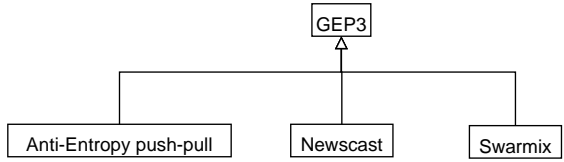}\end{center}

\caption{The GEP3 instances.}

\end{figure}

\textbf{Anti-entropy push-pull \cite{43922}:} this mechanism is targeted
to replicate databases across a network. In this instance of the GEP3
the information collection $S_{v}=C_{v}$ of every $v\in V$ corresponds
to database entries. Every single node $v$ in this setup has a global
view of the network, i.e., $Q_{v}=V$, and the goal is to always maintain
\emph{all} 'freshest' entries. Therefore, the merging and selection
steps are both based on the function \emph{}$t:S\rightarrow\textrm{Time}$,
which associates timestamp values with the database entries.\\
After the nodes $v$ and $w$ have performed one run of Algorithm
\ref{alg:General-epidemic-push-pull-Alg}, both of them end with a
\emph{symmetric} information set, corresponding to all fresh database
entries $C_{v}=C_{w}$. Note that this instance of the GEP3 requires
no neighborhood adaptation.

\textbf{Newscast protocol \cite{jelasity02largescale,newscast03}:}
in this case, news items defined by the protocol are the ones corresponding
to the cache $C$. In contrast to the push-pull mechanism, the peers
in the Newscast protocol have only a partial view of the system. The
size of this network view (or neighborhood) is a fixed number $k$,
which also corresponds to the size of the cache. This is a consequence
of the specification of the news items, which includes information
about the neighbors' IDs. Therefore, the partial view information
is also subject to exchange when executing steps 2 and 3 of the GEP3
(i.e., the epidemic dissemination).

Similar to the push-pull mechanism, the weight function for merging
and selection is also $t:S\rightarrow\textrm{Time}$, but a global
synchronization is not required and just consistency between the timestamps
inside a particular node collection is necessary. The selection function
in this case keeps only the freshest \emph{$k$} news items in the
cache.

When a node $v$ following the Newscast protocol updates not only
its application-dependent information, but also its neighborhood,
then every cycle it gets to know more and more nodes from which
$v$ selects the ones carrying the 'freshest' information.

After a run of the protocol, nodes $v$ and $w$ both end up with
an information set corresponding to the freshest entries in their
caches $C_{v}=C_{w}$ , similar to the push-pull mechanism.

\begin{table*}

\caption{Specific instances of the GEP3\label{tab:Extensions-of-the-GEPPM}}

\medskip{}
\begin{center}\begin{tabular}{|c>{\centering}p{0.1cm}>{\raggedright}p{2cm}|>{\centering}m{2.5cm}|>{\centering}m{2cm}|>{\centering}m{2.1cm}|}
\hline \multicolumn{3}{|c|}{ \textbf{\footnotesize GEP3} }&
\textbf{\footnotesize Anti-entropy push-pull} &
\textbf{\footnotesize
Newscast} & \textbf{\footnotesize Swarmix}\tabularnewline
\cline{4-4} \cline{5-5} \cline{6-6} \hline
{\scriptsize
$C_{v}$} & {\scriptsize ,} & {\scriptsize cache at node}
\emph{\scriptsize v} &
{\scriptsize database entries
\cite{43922}} & \emph{\scriptsize news items} {\scriptsize
\cite{jelasity02largescale}} & {\scriptsize Swarmix
items}\tabularnewline \hline
{\scriptsize
$Q_{v}$} & {\scriptsize ,} &
\parbox[t]{2cm}{
{\scriptsize node} \emph{\scriptsize v}{\scriptsize 's view of the
system }\\
{\scriptsize (i.e.} \emph{\scriptsize v}{\scriptsize 's
neighborhood)} }&
{\scriptsize global view }\\
{\scriptsize ($Q_{v}=V$) }&

{\scriptsize partial view of size} \emph{\scriptsize k} &

{\scriptsize partial view of size} \emph{\scriptsize k}
\tabularnewline \hline

{\scriptsize $u(s)$ ; $s\in C_{v}$} &

{\scriptsize ,} & {\scriptsize utility function}&
\parbox[b]{2.5cm}{
\vspace*{0.1cm}
{\scriptsize $t:S\rightarrow\textrm{Time}$}\\
{\scriptsize (i.e. timestamp)} }&
\parbox[b]{2cm}{
\vspace*{0.1cm}
{\scriptsize $t:S\rightarrow\textrm{Time}$ }\\
{\scriptsize (i.e. timestamp)} }&
\parbox[b]{2.1cm}{
\vspace*{0.1cm}
{\scriptsize $\textrm{similarity}(v,w)$ ; }\\
{\scriptsize $v,w\in V$} }\tabularnewline \hline
\end{tabular}\end{center}

\end{table*}

\section{The Swarmix Epidemic Protocol\label{sec:The-Swarmix-Epidemic-Protocol}}

Our Swarmix epidemic protocol is based on the framework provided by
the GEP3. In this section we detail the extensions of the model, and
its specific characteristics (Figure \ref{tab:Extensions-of-the-GEPPM}).

Our protocol supposes a collection of \emph{Swarmix items} called
\emph{cache}, corresponding to the set $C_{v}$ in the GEP3. Each
peer $v$ in the overlay network locally stores its own cache, which
has a limited number $k\in\mathbb{N}$ of \emph{cache entries}.

\begin{figure}
\begin{center}\includegraphics[%
 scale=0.2]{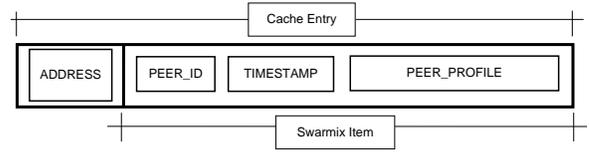}\end{center}

\caption{\label{fig:swarmix-item}The format of Swarmix items and
cache entries.}

\end{figure}

The format of a Swarmix item stored at peer $v$ is depicted in Figure
\ref{fig:swarmix-item}. Suppose that the information contained in
$v$'s Swarmix item belongs to its neighbor $w$. This information
includes \emph{}$w$\emph{'}s unique ID (e.g., its pseudonym), a vector
representation of its profile (e.g., its ratings), and a timestamp
corresponding to the Swarmix item creation event. Note that, as in
Newscast, no global synchronization is needed, just consistency between
the timestamps inside a particular peer cache is required.

Note that \emph{$k$} also corresponds to $v$'s neighborhood size,
i.e., to the size of set $Q_{v}$ specified by the GEP3, which corresponds
to $v$'s {}``view'' of the network. Obviously, this is a partial
view, but a dynamic one, constantly changing as the protocol is executed.

The Swarmix item is also associated to $w$'s network address, which
is used by \emph{}$v$ \emph{}to contact $w$ when performing the
protocol. This extra information, plus the Swarmix item itself, constitutes
what we called a \emph{cache entry}.

Note that the partial network view $Q_{v}$ of peer \emph{}$v$ \emph{}is
also subject to information exchange when executing the push-pull
steps, which in terms of GEP3, implies that the data ID coincides
with the data source (i.e., $\textrm{id}=\textrm{src}$).

The GEP3 utility function $u$ is defined to be the rating profile
similarity function:

\[
u_{v}(s)\textrm{ := sim$(v,\textrm{id}(s))$}\]

which computes the similarity between peers $v$ and $w$ based on
their respective rating profiles, e.g., cosine similarity.

The utility function assigns the similarity weight to each Swarmix
item $s$ in $v$'s cache $C_{v}$, and permits to establish a total
order over $v$'s cache entries (i.e., $>_{s\in C_{v}}$), which \emph{}constitutes
the selection criteria to decide what Swarmix items are to be kept
as the {}``best'' ones, when performing the selection step of the
GEP3.

We merge peers' $v$ \emph{}and $w$ \emph{}caches by retaining the
most useful version of each Swarmix item:

\[
\text{merge}(C_{v},C_{w}):=\{\textrm{argmax}_{\substack{r\in C_{v}\cup C_{w},\\
id(r)=\text{id}(s)}
}u(r)\:|\: s\in C_{v}\cup C_{w}\}\]

As selection function as well as for neighborhood selection we
opted for retaining a fixed number $k$ of most useful peers as
described above already. As the size of the cache and the size of
the neighborhood are the same, the neighbors are just the peers
specified by the Swarmix items in the updated cache after one
round of the protocol.

\begin{description}
\item [Observations]~
\end{description}
\begin{enumerate}
\item The entries in $v$'s and $w$'s cache after one execution of
the protocol are in general not the same (i.e., the Swarmix
protocol is \textbf{asymmetric}), different to both of the
protocols discussed before, i.e., anti-entropy push-pull and
Newscast. Note that the asymmetry of caches and neighborhoods is
not simply a consequence of an asymmetric similarity measure. In
fact, cosine similarity itself is symmetric. \item The global
communication cost of one cycle for the overall system depends on
the cache size $k$, and the size of the user rating profile. In a
cycle, each peer initiates exactly one information exchange
session which involves the transfer of at most $2k$ cache entries.
The size of the cache can be seen from Figure
\ref{fig:swarmix-item}. Clearly, the global communication costs of
one cycle grow linearly with the
network size, but are evenly distributed over the set of peers.\\
For the communication costs for a single peer, Jelasity et al.
\cite{jelasity02largescale} prove that the distribution of the
incoming connections is close to a Poisson distribution with
$\lambda=1$. Their proof is based on the random pickup step of the
their Newscast epidemic algorithm, assuming unbiased random cache
content, and for a large network size. Our protocol shares the
same basic random pickup step, and under the same assumptions, we
can expect that the same distribution of incoming connections
applies to it. \item From the point of view of \emph{scalability}
the sharing of communication cost over time, and among all peers
is an important advantage: independent of the system size, each
peer will experience the same predictable load without peaks.
\item Peers joining the network do not require any special
sequence of communications, the new peer simply has to initialize
its cache with at least one known peer (e.g., a \emph{buddy})
which is already on-line, and start to execute the protocol. \item
A peer voluntarily leaving the overlay simply has to stop
communicating, i.e., leavings are treated as failures.
\end{enumerate}

\section{Recommendation Algorithm\label{sec:Recommendation-Algorithm}}

The problem space of automated collaborative filtering can be formulated
as a matrix \emph{R} of users versus items. Each cell of the matrix
\emph{R} represents a user's \emph{rating} on a specific item, and
each row corresponds to a \emph{user profile}. The task of the recommender,
under this formulation, is to predict values for specific, empty cells;
i.e., to predict a user's rating for a not-yet-rated item.

A neighborhood-based collaborative filtering recommender system comprises
the three fundamental steps described by Herlocker et al. \cite{Herlocker2002EmpiricalAnalysisDesignCFAlgs}:

\begin{enumerate}
\item \textbf{Similarity computation}. Compute similarity of all users'
profiles with the profile of the target user (i.e., the one requesting
recommendations, also called active user).
\item \textbf{Neighborhood Formation}. Select a subset of users as a set
of predictors: These neighbors correspond to the $c$ most similar
users for the active user.
\item \textbf{Aggregation and prediction computation}. The active user's
profiles ~are ~aggregated ~computing ~the\\
union of consumed items. The system also removes items already consumed
by the active user, in order to guarantee that just new items are
recommended. A weight is associated to each item based on its importance
in the aggregation; consequently, the best $N$ items, having the
highest weights, are reported to the active user as the final recommendations.
\end{enumerate}
These steps may overlap or might be slightly different depending on
the specific recommender system that implements the algorithm.

In case of a recommender system operating in a standard client/server
architecture, these steps are performed by the server in a centralized
manner, while the role of the clients is minimal and limited to providing
access to the final recommendation lists.

The Swarmix architecture distributes among all participants the execution
of these three tasks. The neighborhood formation is provided by the
Swarmix epidemic protocol, where every peer is associated with a neighborhood
of like-minded peers.

This neighborhood corresponds to the partial view of the network (i.e.,
the cache) as discussed in the previous section. While the protocol
executes, peer $v$'s neighborhood changes continuously, selecting
in every run the \emph{$k$} most similar peers.

The similarity computation is also part of our epidemic protocol,
and is performed by each peer in the overlay network. We perform
the similarity computation using the \emph{cosine-based} measure:

\[
\textrm{sim}(v,w):=\cos(v,w):=\frac{\langle v,w\rangle}{||v||\cdot||w||}\]

where by abuse of notation $v$ and $w$ denote the respective rating
profiles of peers $v$ and $w$. Alternatively, any other similarity
measure proposed in the literature could be used, e.g., Pearson correlation,
Spearman rank, etc.

Finally, each peer is able to compute its recommendation list based
on its neighborhood, that is, through its cache entries. For our architecture,
we have implemented the \emph{most-frequent items} approach suggested
by Sarwar et al. \cite{DBLP:conf/sigecom/SarwarKKR00}. Their technique
can be seen as a majority voting election scheme, were each of the
members of peer $v$'s neighborhood casts a vote for each of the items
he has consumed. Those \emph{N} items with most votes, and new to
the active user, are the recommendations.

\section{Experiment Outline\label{sec:Experimental-Setup-and-Metric}}

To evaluate the result of the top-\emph{N} (with \emph{N}=10)
recommendations provided by our distributed architecture, we split
the dataset into \emph{training} and \emph{test} set by randomly
selecting a single rating (a \emph{hidden} item) for each user to
be part of the test set, and used the remaining ratings for
training. Breese et al. \cite{DBLP:conf/uai/BreeseHK98} called
this kind of experimental setup \emph{all-but-1} protocol.

The nearest neighbors and top-10 recommendations were computed using
the training set only.

The quality was measured by looking at the number of \emph{hits},
which corresponds to the number of items in the test set that were
also present in the top-\emph{N} recommended items returned for each
peer. More formally, \emph{hit}-\emph{rate}, is defined as

\[
\textrm{hit-rate}:=\frac{\sum_{v\in V}hit(v)}{|V|}\]

where \emph{hit(v)} is a binary function that returns 1, if the hidden
item is present in $v$'s top-\emph{N} list of recommendations, and
0 otherwise. A \emph{hit-rate} value of 1.0 indicates that the system
was able to always recommend the hidden item, whereas a \emph{hit-rate}
of 0.0 indicates that the system was not able to recommend any of
the hidden items.

In order to ensure that our results are not sensitive to the particular
training-test partitioning of each dataset, we performed five different
runs for each of the experiments, each time using a different random
partitioning into training and test sets. The results reported in
the rest of this section are the averages over these five trials.
Furthermore, to better compare the results we also present the confidence
intervals for the mean estimation at a 95\% confidence coefficient,
when appropriate.

For the time analysis, 100 simulation cycles were performed. In each
cycle, every peer $v$ initiates a Swarmix communication session
and, immediately after finishing, $v$'s recommendations are computed.
Thus, in each simulation cycle \emph{n} Swarmix sessions are performed,
and each peer $v$ receives its top\emph{-}10 recommendations once
during the cycle.

In all experiments, a \emph{cache} value of size 20 was used when
executing the Swarmix protocol. This corresponds to the typical
neighborhood size adopted for generic recommender systems.

The \emph{bootstrapping procedure} for the P2P network is as
follows. We start with a connected network, but with an initially
unbalanced neighborhood structure with an average neighborhood
size of 2.494 neighbors per peer. Therefore, none of the peers has
its cache filled. This network state corresponds to simulation
cycle 0, i.e., before starting to execute the Swarmix protocol.

The recommender system was also deployed in a centralized architecture,
as a reference for the performance of our distributed architecture.

\section{Experimental Results\label{sec:Experimental-Results}}

We evaluate the performance of the recommender system implemented
using a 88\% subset of the original 'small' MovieLens dataset\cite{MovieLensDataset2004}
containing 943 users which have issued 88,263 explicit ratings, on
a scale from 1 to 5, for 1,457 movies.%
\footnote{ Actually we dropped all ratings of movies that could
not be identified in the Amazon taxonomy by Ziegler et al.
\cite{1031252}, as we plan to compare with taxonomy-based
recommendation strategies in upcoming work.
} Each of the users has issued at least 15, and on average 93.6 ratings.

A user in the dataset becomes a \emph{peer} in our simulation environment.

\begin{figure*}[t]
\begin{center}\begin{tabular}{cc}

\includegraphics[%
 scale=0.65]{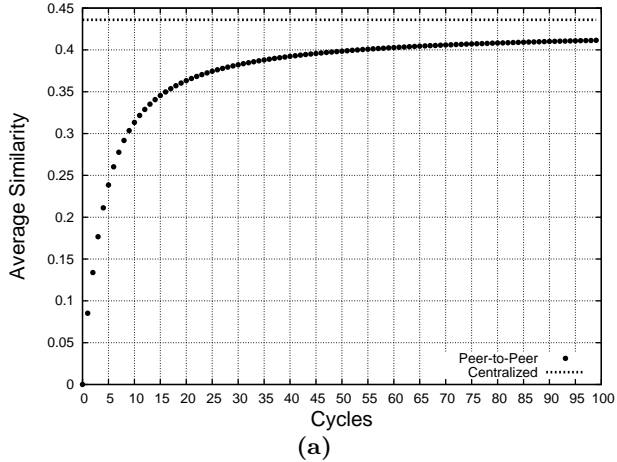}
&

\includegraphics[%
 scale=0.65]{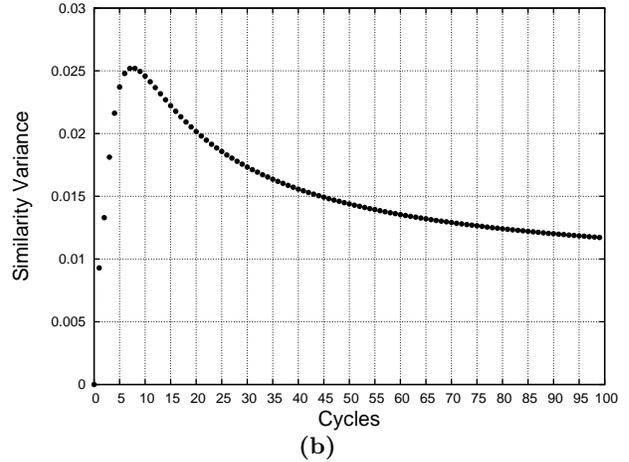}
\tabularnewline
\textbf{(a)}&
\textbf{(b)}\tabularnewline
\end{tabular}\end{center}

\caption{(a) Average similarity, (b) similarity variance.
\label{fig:Average-Similarity-And-Variance(CF)}}

\end{figure*}

\subsection{Neighborhood Quality}

To evaluate the quality of the neighborhoods formed and maintained
by the Swarmix epidemic protocol, we focused our attention on the
average similarity score between each peer and its neighbors. For
each of the 100 simulation cycles, Figure \ref{fig:Average-Similarity-And-Variance(CF)}(a)
shows the similarity values averaged over all peers in the overlay.
The figure also presents, as reference, the average similarity values
in the case of a centralized architecture, where each peer $v$ is
associated with its 20 most similar neighbors. Obviously, the similarity
values for the centralized architecture remain constant throughout
the simulation cycles, and they represent an upper bound for the performance
of our distributed architecture.

The Swarmix protocol is able to maintain neighborhoods whose average
similarity improves on every cycle, converging to values close to
the ones in a centralized architecture. Furthermore, the variance
of the average similarity scores decreases over time (Figure \ref{fig:Average-Similarity-And-Variance(CF)}(b)).
Note that the initial behavior is explained by the bootstrapping procedure
used, which initializes a peer's cache not at full capacity, but with
just 2 or 3 entries maximum. This makes the variance worse during
the initial cycles, while filling the caches, but then starts improving
to smaller values throughout the rest of the simulation.

Note that even though we have applied \emph{significance weighting}
to improve recommendation quality, as recommended by Herlocker et
al. \cite{Herlocker2002EmpiricalAnalysisDesignCFAlgs}, the similarity
values reported here correspond to the ones \emph{before} the application
of such weighting. The reason is that we want to first assess the
performance of the protocol, comparing it with respect to the expected
behavior of finding more and more similar peers during its continuous
execution. Significance weighting is assumed for all experiments involving
the computation of recommendations.

\subsection{Recommendation Quality}

Next, we look at the \emph{hit-rate} score, which help us evaluate
whether the system is making recommendations for items that the peers
will recognize and value.

\begin{figure}[t]

\begin{center}\includegraphics[%
 scale=0.65]{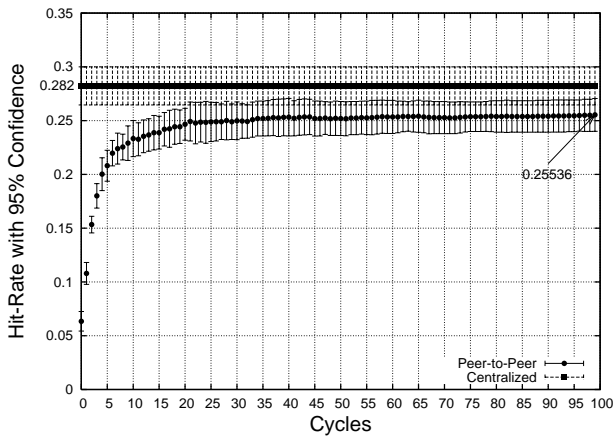}\end{center}

\caption{Hit-Rate\label{fig:Hit-Rate-pure-CF}}

\end{figure}

The hit-rate for the pure-CF recommender implementation is presented
in Figure \ref{fig:Hit-Rate-pure-CF}.

In looking at the figure one can observe how the recommendation quality
improves over time, as a consequence of the intra-neighborhood similarity
improvement. The series shows that for the Swarmix architecture, the
\emph{hit-rate} measure is nearly equal to the central server's.

\begin{figure}

\begin{center}\includegraphics[%
 scale=0.65]{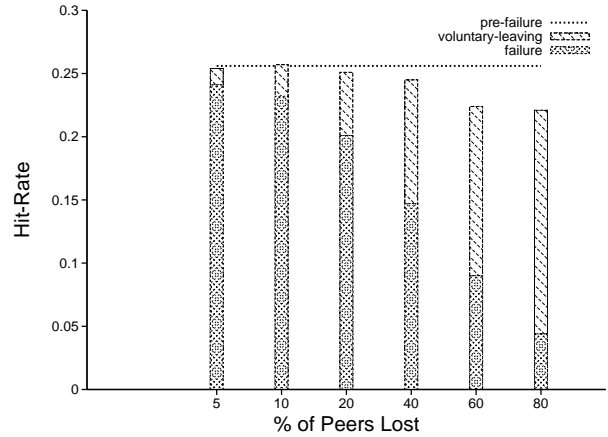}\end{center}

\caption{Failure and voluntary-leaving
effect.\label{fig:Failure-and-voluntary-leaving}}

\end{figure}

\subsection{Failures and Voluntary-Leavings}

We analyze the effect of peer failures and voluntary leavings from
the overlay. For these experiments we applied a disturbance at cycle
50 and let the Swarmix network evolve. Finally, we report the \emph{hit-rate}
values corresponding to the ones at cycle 100. Peer losses of 5\%,
10\%, 20\%, 40\%, 60\% and 80\% were simulated.

\textbf{Failures.} We perform these experiments considering that a
peer $v$, disconnected from the network as a consequence of a failure,
is not able to receive recommendations, but still wants to receive
them. Therefore, we consider the total number of peers (i.e., 943)
when computing the \emph{hit-rate}.

\textbf{Voluntary leavings.} In case of a peer leaving the network
voluntarily, we modified the \emph{hit-rate} to take into consideration
only those peers that remain connected to the overlay. If \emph{L}
represents the set of peers that have left the network, the hit-rate
for voluntary leavings is computed as\[
\textrm{hit-rate}_{\textrm{voluntary\textrm{-}leavings}}:=\frac{\sum_{v\in(V\backslash L)}hit(v)}{|V\backslash L|}\]
Therefore, we assumed that peers leaving the network do not want
to receive their recommendations anymore. Note that this is a worst
case scenario, because they are able to receive recommendations, locally
computed from the cache entries, even in the case when no connection
to the overlay exists (i.e., using their cache entries).

Figure \ref{fig:Failure-and-voluntary-leaving} shows the simulation
results.

\subsection{Network Characteristics}

\begin{figure*}
\begin{center}\begin{tabular}{cc}

\includegraphics[%
 scale=0.6]{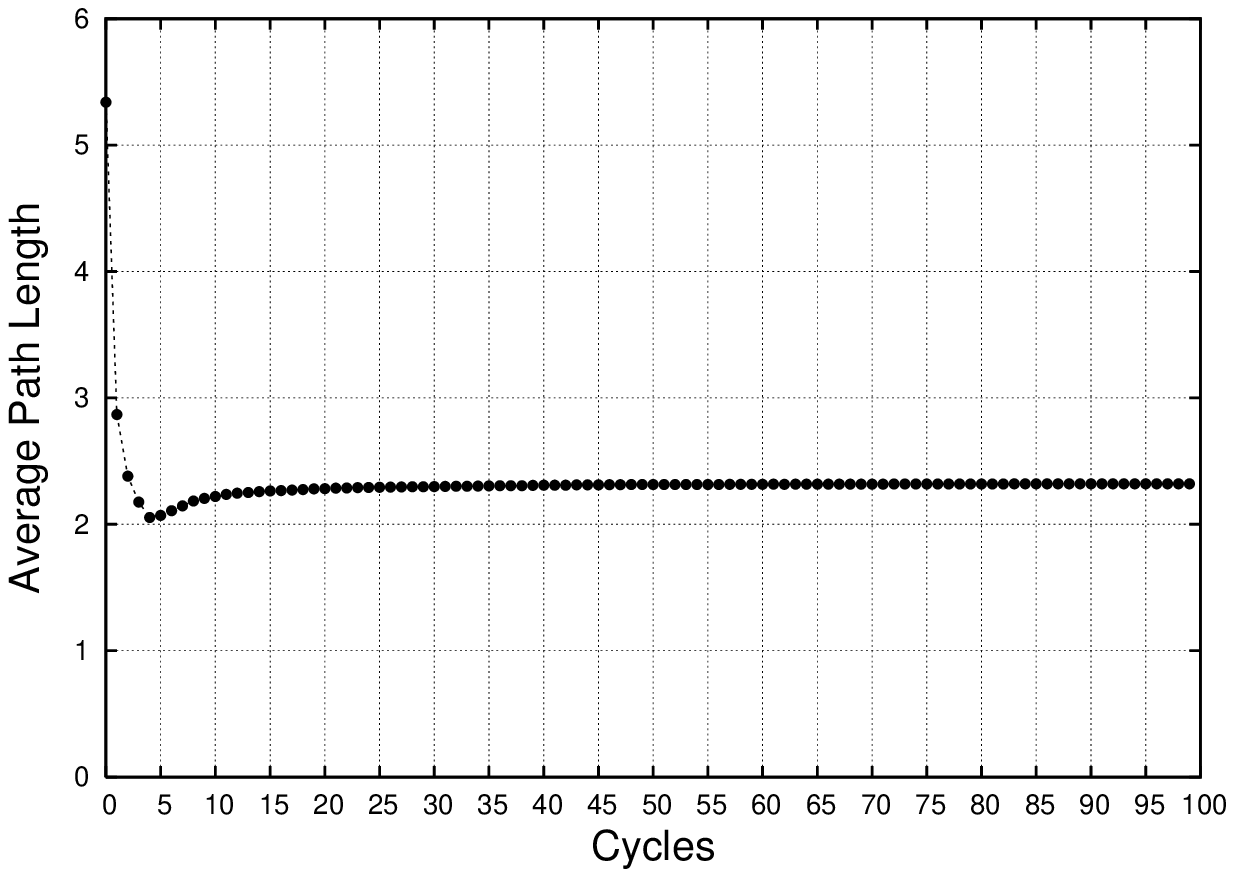}
&
\includegraphics[%
 scale=0.6]{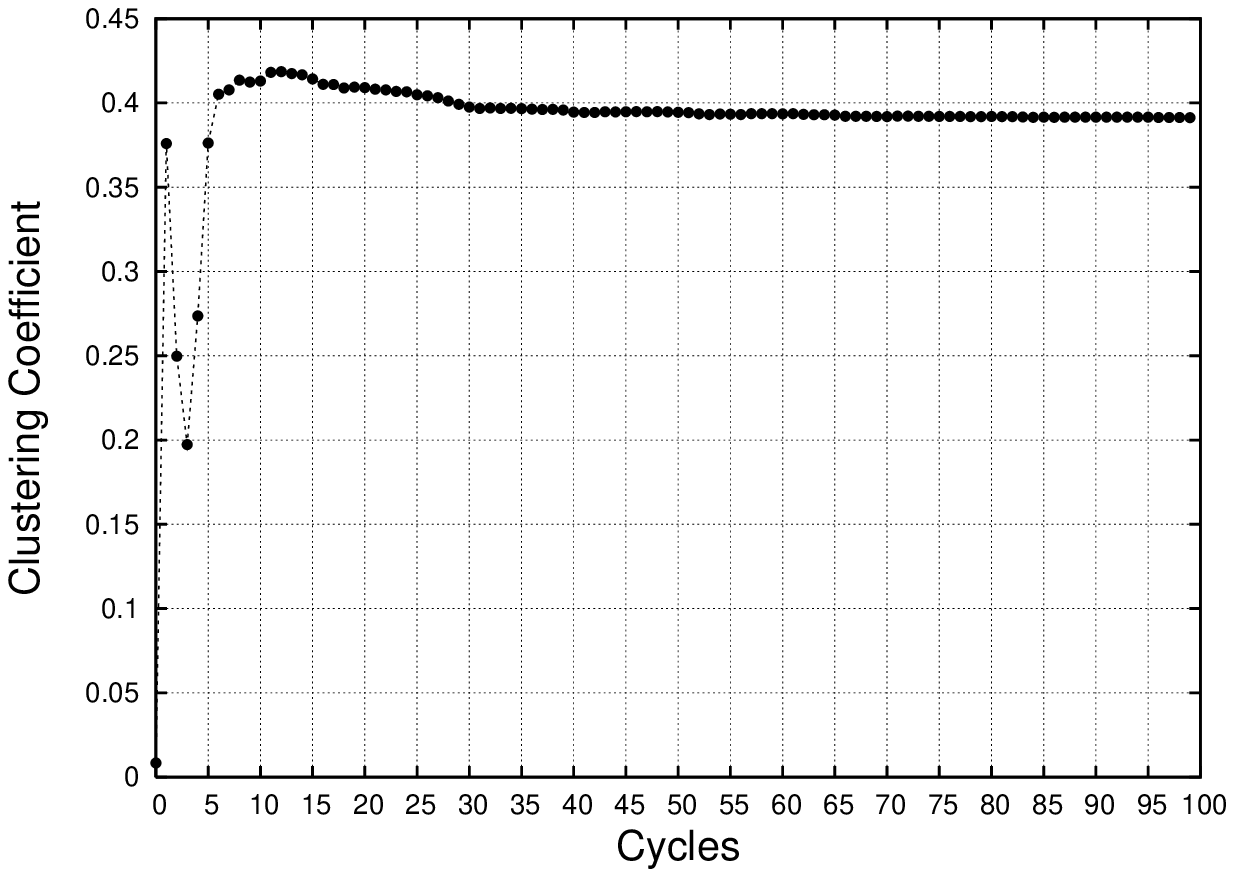}\tabularnewline
\textbf{(a)}&
\textbf{(b)}\tabularnewline
\end{tabular}\end{center}

\caption{(a) Average path length, (b) clustering
coefficient.\label{fig:Avg-Path-Length-AND-Clust-CF}}

\end{figure*}

Complex networks are frequently used to model a wide variety of systems
of high technological, social, biological and intellectual importance.
Ordinarily, the connection topology of such network models is assumed
to be either completely regular or completely random, but many social,
biological and technological network systems lie somewhere in between
these two extremes. Watts and Strogatz \cite{WS98} dubbed these network
systems as \emph{small-world} networks, in analogy with the small
world phenomenon \cite{milgram67smallworld}.

The small-world property appears to characterize most complex networks,
such as electrical power grids, neural networks, science collaboration
graphs and so forth \cite{WS98,DBLP:journals/corr/cond-mat-0106096}.

Two characteristics set small-world networks apart: first, a
\emph{small} \emph{average path length}, typical of random graphs,
defined as the number of edges in the shortest path between two
vertices, averaged over all pairs of vertices; and it measures the
typical separation between two vertices in the graph (a
\emph{global} property). Second, a \emph{large clustering
coefficient,} independent of network size, measures the
cliquishness of a typical neighborhood (a \emph{local} property),
(i.e., how many of a node's neighbors are connected to each
other).

We can observe from Figure \ref{fig:Avg-Path-Length-AND-Clust-CF}(a)
that very low average path lengths are obtained. Note that in order
to get a finite value, the network has to be connected.

The \emph{average clustering coefficient}, taken over all nodes in
the overlay, is shown in Figure
\ref{fig:Avg-Path-Length-AND-Clust-CF}(b). Observe that the values
shown are relatively high (if our networks were random, the
clustering coefficient would be expected to be approximately
$(k/n)=(20/943)=0.02121$
\cite{WS98,DBLP:journals/corr/cond-mat-0106096}).

The behavior during the initial cycles is explained by the
application of the bootstrapping procedure previously explained.
However, few cycles after starting to perform the Swarmix epidemic
protocol, the average path length and clustering coefficient
values converge quickly to their eventual values.

The small average path length values, in combination with the
relatively high clustering coefficients, allow us to conclude that
our Swarmix P2P overlay is a \emph{small-world} network. This
means that information dissemination is efficient, for two
arbitrary nodes are separated by a few links only. Furthermore, it
is expected that those high clustering coefficient values would
provide certain resilience for network partitioning, e.g. in the
presence of peer failures.

Note that these properties are not maintained explicitly, but they
\emph{emerge} from the underlying simple epidemic-style Swarmix protocol.

\section{Related Work\label{sec:Related-Work}}

In this section, we present some examples of related research on
deploying recommender systems in distributed architectures.

PocketLens \cite{PocketLens2004} is a P2P-based collaborative
filtering algorithm that incrementally updates an item-item model
\cite{963776} for later use to make recommendations. In contrast
to PocketLens, Swarmix builds a \emph{user-based} matrix
\cite{Herlocker2002EmpiricalAnalysisDesignCFAlgs} for each peer
$v$, where the users in the matrix correspond to $v$'s neighbors
only, avoiding scalability problems when the amount of users in
the network increases.

Haase et al.~\cite{hasse2005usgDrvEvolPersonalOnto} deploy a CF
recommender system over a P2P-based personal bibliography
management tool. The recommender system assists users in the
management and evolution of their personal ontology by providing
detailed suggestions of ontology changes. These suggestions are
based on the usage information of the individual ontologies across
the P2P network. Swarmix is domain-independent and could be tuned
to deliver recommendations of actions, not only items, only
requiring a meaningful way to represent user profiles in order to
compute their similarity for neighborhood formation.

An entirely distributed CF algorithm called PipeCF, based on a
content-addressable distributed hash table (DHT) infrastructure,
is presented in \cite{PengHaletal2004}. Swarmix depends on a
epidemic-style protocol for information dissemination.

One area of research that intersects with peer-to-peer recommender
systems systems is that of mobile and intelligent software agents.
Yenta \cite{267732}, for example, is a decentralized multi-agent
system that focuses on the issue of finding other peers with similar
interests using referrals from other agents.

\section{Conclusion and Future Work\label{sec:Conclusion-and-Future}}

In this paper, we presented a distributed architecture for recommender
systems, based upon epidemic-style protocols and geared towards peer-to-peer
scenarios such as the Semantic Web, the Grid, and so forth. Our main
focus in marrying these two worlds, i.e., the client/server-dominated
world of recommender systems and the emerging peer-to-peer paradigm,
into one coherent model has been to preserve the benefits of both,
namely recommendation quality and scalability, robustness, and resilience
to failure.

To this end, we empirically evaluated our architecture, assessing
neighborhood similarity and the quality of recommendations first.
Our results showed that our distributed architecture is able to maintain
neighborhoods whose intra-similarity steadily improves over time.
Final recommendations obtained are comparable to the ones expected
in a centralized architecture, an implicit consequence of these afore-mentioned
high-quality neighborhoods.

We also investigated characteristics of the evolving network, observing
the emergence of a small-world network from its initial random-graph
topology \cite{ER59}. Consequently, we derive that information dissemination
over the Swarmix network is efficient, for any two nodes are separated
by small geodesic distances only. Moreover, we expect those relatively
high clustering coefficients $C(p)$ to provide substantial resilience
against network partitioning, e.g., in case of peer failures. We also
examined the network behavior in the presence of voluntary leavings,
and failures.

Our research breaks new grounds and enters territory that still appears
largely untouched. Hence, there remain numerous branches for future
research to take.

First, owing to the very nature of the epidemic protocol combined
with our \emph{similarity}-based weighting function, our current setup
risks the perpetual dissemination of outdated profile information.
Assigning timestamps to profiles and incorporating an {}``aging''
mechanism solves this issue in an efficient and simple manner. Furthermore,
we also have to consider the dynamics of user preferences over time.

Second, we would like to better understand and characterize the social
network characteristics of the emerging web. We already know about
its small-world traits, i.e., high clustering coefficient, and small
path lengths between any two peers \cite{WS98}. We are now interested
in its evolutionary \emph{dynamics}, i.e., investigating the presence
or absence of typical {}``rich get richer phenomena'' \cite{BA99},
leading to power-law distributions of node degrees.

Other, more practical and user-centric considerations include privacy
and security issues. Though being key motivations for crafting decentralized
recommender systems, they have not been addressed in this paper. To
this end, we believe that computational trust and trust propagation
models \cite{ZL04} inherently solve the issue of privacy, and security
in particular, and could likewise serve as efficient and scalable
means for neighborhood selection.

We envision our architecture as part of the Semantic Web's infrastructure,
with peers being represented by Semantic Web agents. Owing to different
domains that users are interested in, e.g., not only books but also
music, movies, etc., several instances of the protocol are supposed
to run simultaneously in one single agent, allowing the computation
of high-quality recommendations for each of these domains of interest.

\end{document}